\documentclass[letterpaper, 10 pt, conference]{ieeeconf}  

\IEEEoverridecommandlockouts                              

\usepackage{graphics} 
\usepackage{epsfig} 
\usepackage{mathptmx} 
\usepackage{times} 
\usepackage{amsmath} 
\usepackage{amssymb}  
\usepackage{graphicx}
\usepackage{makecell}
\usepackage{caption}
\usepackage{mathrsfs}
\usepackage[ruled,vlined]{algorithm2e}
\usepackage[font={small}]{caption}
\usepackage[mathscr]{euscript}
\usepackage{bm}
\usepackage{pifont}
\usepackage{siunitx}

\usepackage{subcaption}

\usepackage{array}
\newcommand{\PreserveBackslash}[1]{\let\temp=\\#1\let\\=\temp}

\newcolumntype{C}[1]{>{\PreserveBackslash\centering}p{#1}}
\newcolumntype{R}[1]{>{\PreserveBackslash\raggedleft}p{#1}}
\newcolumntype{L}[1]{>{\PreserveBackslash\raggedright}p{#1}}

\newcommand{\doubleBlind}[2]{#2}

\DeclareMathOperator*{\argmin}{arg\,min} 

\usepackage[hidelinks]{hyperref}
\hypersetup{
    colorlinks=true,
    linkcolor=blue,
    urlcolor=blue,
    citecolor=blue
}

\usepackage{balance}

\usepackage[utf8]{inputenc}
\usepackage[english]{babel}
\usepackage[backend=biber,sorting=none,citestyle=numeric-comp,doi=false, url=false,isbn=false]{biblatex}

\title{\LARGE 
Reactive Humanoid Multi-Contact Using Learned Stability Models
}

\author{\doubleBlind{Authors ommitted}{Stephen McCrory$^{1,2}$, Beomyeong Park$^{1}$, Nicholas Kitchel$^{1}$, Nehar Poddar$^{1,2}$, Robert Griffin$^{1,2}$}
\thanks{$^{1}$\doubleBlind{Ommitted}{Author is with the Florida Institute for Human and Machine Cognition.}}%
\thanks{$^{2}$\doubleBlind{Ommitted}{Author is with the University of West Florida.}}%
\thanks{\doubleBlind{Ommitted}{This work was supported through ONR Grant No. N00014-19-1-2023. Email: \url{smccrory@ihmc.org}}}%
}

\begin{document}
\maketitle
\thispagestyle{empty}
\pagestyle{empty}

\begin{abstract}

We present a planning and control approach to reactively use hand contacts to stabilize a humanoid in low stability scenarios, where only using feet contacts may result in a fall. Candidate contacts are sampled within the robot's reachable workspace, and a preview is computed by rolling out the centroidal dynamics through pre-impact, impact and post-impact phases. Sampled points are scored based on the Center of Pressure (CoP) control authority at the post-impact phase. Central to our approach is a learned model of the robot's CoP region during post-impact, which enables rapid evaluation of candidate contact points compared to traditional optimization-based methods. The presented planner has two stages: the first selects an optimal bracing region and the second computes an optimal bracing point within the region. Our simulation results demonstrate an average increase in impulse resilience of 89\% over recovery without hand contacts and 17\% over a naive planning strategy (closest reachable region). We validate our framework on hardware, performing push tests while standing and walking. The standing trials show an average 43\% reduction in stabilization time compared to naive hand placement and the walking trials demonstrate a 18\% reduction compared to baseline recovery (without hand contacts).

\end{abstract}

\section{Introduction}


As humanoid robots become capable of performing tasks in real-world scenarios, reliability becomes increasingly important. For this reason, push recovery has a rich history in legged robotics, with the broad goal of maximizing robustness \cite{kajita1991study, pratt2006capture, del_prete_zero_2018, lin2020robust}. However, in spite of numerous successful strategies for bipedal push recovery, existing approaches are largely restricted to quasi-flat terrain and feet contacts. Making full use of the humanoid form-factor means utilizing upper body contacts to augment lower-body recovery techniques. Additionally, it requires models that go beyond quasi-flat terrain and handle arbitrary contact surfaces.

\begin{figure}[t]
    \centering
    \includegraphics[width=\columnwidth]{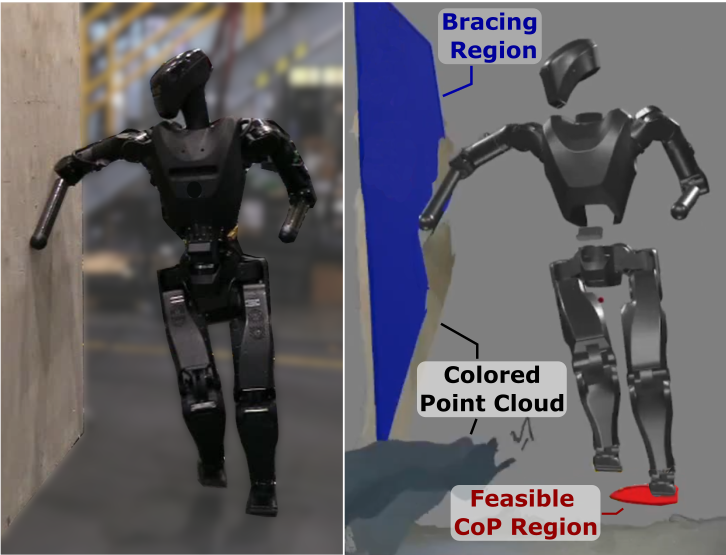}
    \caption{(left) \doubleBlind{The robot}{Alex} reactively bracing against a wall to stabilize after being pushed while walking. (right) The robot's head-mounted depth sensor detects the bracing surface (blue) and the robot's feasible CoP region (red) is previewed to select a bracing point.}
    \label{fig:title_figure}
    \vspace{-4mm}
\end{figure}

In this work, we propose a sample-based hand contact planner that reactively braces in low stability scenarios where the use of only foot contacts may result in a fall, such as the scenario in Fig. \ref{fig:title_figure}. Sampled contact points are previewed by rolling out the robot's centroidal dynamics through pre-impact, impact and post-impact phases. The pre-impact and impact phases can be rolled out using conventional closed-form models such as the Linear Inverted Pendulum Model (LIPM) \cite{kajita1991study}. In contrast, the post-impact contact state cannot be modeled as a basic LIPM, but instead using optimization-based models that are compatible with arbitrary contact sets \cite{bretl_testing_2008, orsolino_feasible_2020}. Each roll-out is scored based on the Center of Pressure (CoP) control authority at the post-impact phase.

A central part of our approach is a set of neural networks that predict the feasible CoP placement at the post-impact rollout phase. We train these models using reference CoP data computed through conventional optimization-based techniques. These networks enable rapid evaluation of the robot's post-impact state, with around 40$\times$ speed-up compared to optimization-based techniques. We apply these networks in a contact planner which performs two passes for each hand. The first pass selects the optimal bracing surface by scoring the closest reachable point on each candidate surface. The second pass samples within the optimal surface (if one exists) to compute the optimal bracing point. 


\subsection{Related Work}


Reactive contact planning for legged robots has been widely studied, with many successful hardware demonstrations. Reduced-order models such as the Linear Inverted Pendulum Model \cite{kajita1991study}, the Angular Linear Inverted Pendulum \cite{gong2021one} and Single Rigid Body Model \cite{kim2024reduced} enable efficient previews of the robot's centroidal dynamics. In turn, these models have been leveraged in heuristics for step placement, such as Raibert's heuristic \cite{raibert1986legged} and the Capture Point \cite{pratt2006capture, englsberger_3d_2015}. Since these heuristics have closed-form linear representations, they have been applied in mixed-integer frameworks that select optimal contact surfaces online \cite{acosta2025perceptive, kim2025model, tonneau_sl1m_2020, corberes2025perceptive}. Following this approach, Mason et al. \cite{mason18mpc} reactively plan hand contacts by first previewing a footstep sequence containing a slack variable for centroidal tracking. If this slack variable is sufficiently high, a solver activates which scores candidate hand contacts based on a recovery preview, given a heuristic maximum hand pushing force. Such applications of centroidal models have been successful, but have the caveat of requiring linear or quadratic functions to encode cost and constraints and often are applied only to quasi-flat terrains.

A related line of research addresses motion feasibility under arbitrary contact conditions given unilateral, friction and actuation constraints. Seminal work by Bretl et al. \cite{bretl_testing_2008} posed this as a Linear Program, enabling efficient quasi-static stability assessment. This approach has been extended towards improving solve times \cite{del_prete_fast_2016}, addressing dynamic stability \cite{audren_3-d_2018} and modeling actuation constraints \cite{orsolino_feasible_2020}. Marucci et al. apply a similar model to multi-contact planning on a piecewise affine map \cite{marcucci2017approximate}. Other centroidal-based reactive recovery strategies have explored adapting gait frequency and posture to achieve high stability \cite{li2025adapting, ferrolho2023roloma}.

There have also been machine-learning based approaches that aim to model stability and assess contact sequences. One approach, and the one taken in this work, is training a network to mimic an expensive optimization-based stability model \cite{orsolino_rapid_2021}. Alternatively, Lin et al. use simulation data to train a set of networks to predict zero- and one-step capturability, given a robot's state and environment \cite{lin2020robust}. These networks are applied in a graph-search that optimizes for predicted success in the presence of disturbances, such that the robot walks nearby potential bracing surfaces. This approach of using procedurally generated simulation environments has seen many applications in legged motion planning \cite{chavez2018learning, miki2024learning}. Rouxel et al. use datasets from teleoperated hardware experiments to train a network to load/unload a humanoid's hand based on predicted stability \cite{rouxel2024flow}. 




\subsection{Contributions}

This work builds on the core concepts of reduced-order models to achieve reflexive response times, while maintaining accurate stability modeling under arbitrary contact conditions by applying learned stability models. Our work particularly builds on the MPC-style bracing planner by Mason et al. \cite{mason18mpc} and the concept of optimal region selection by Corbères et al. \cite{corberes2025perceptive}. In building on these approaches, this work contains the following novel contributions.

\begin{itemize}
 \item A reactive ($\leq $10\,ms) contact planner that models recovery for arbitrary (e.g. tilted or vertical) surfaces.
 \item A set of neural network models that predict the robot's feasible CoP placement given arbitrary combinations of hand and feet contacts.
 \item Simulation demonstrations of the planner's capability to out-perform a naive heuristic through a comparison of the minimum unrejectable impulse.
 \item Hardware demonstrations of the planner and controller's capacity to use hand contacts to recover from high impulses.
\end{itemize}

\section{Background}

We review a family of optimization problems that model the stability of a legged robot which has a set of arbitrary unilateral- and friction-constrained contacts. The robot's stability is modeled through the optimization problem below.

\begin{equation} \label{eq:ContactFeasibility}
\begin{aligned}
    \max_{\mathbf{c},\mathbf{f}} \qquad & \mathbf{a}_1^T\mathbf{c} + \mathbf{a}_2^T\mathbf{f} \\
    \textrm{s.t.} \qquad & \mathbf{A}_1\mathbf{c} + \mathbf{A}_2 \mathbf{f} = \mathbf{t}(\mathbf{g}) \\ 
    & \mathbf{f} \in \mathscr{K} \\
    & \boldsymbol{\tau}^{-} \leq \mathbf{h}(\mathbf{q},\dot{\mathbf{q}}) -\mathbf{J}(\mathbf{q})^T\mathbf{f} \leq \boldsymbol{\tau}^+, \\  
\end{aligned}
\end{equation}
where the robot has $n$ total degrees of freedom, $n_j$ actuated joints, $n_c$ contact points, $\mathbf{q} \in \mathbb{R}^n$ is the whole-body configuration vector, $\mathbf{c}\in\mathbb{R}^3$ is the robot's CoM, $\mathbf{f}\in\mathbb{R}^{3n_c}$ is a stacked vector of reaction forces, $\mathbf{g}\in \mathbb{R}^3$ is the gravity vector, $\mathbf{J}_c \in \mathbb{R}^{n_j \times 3n_c}$ is a stacked matrix of contact Jacobians, $\boldsymbol{\tau}^{-/+}$ are the min/max joint torque bounds, $\mathbf{h}$ designates gravity and coriolis bias torques and $\mathscr{K}$ is the stacked set of friction cones. Prior works in contact planning focus on quasi-static ($\mathbf{a}_1\neq\mathbf{0}$, $\mathbf{a_2}=\mathbf{0}$) implementations of this problem \cite{bretl_testing_2008, del_prete_fast_2016, orsolino_feasible_2020}. Other works apply this towards dynamic stability, such as Audren et al. \cite{audren_3-d_2018} in which a convex polyhedron of feasible CoM accelerations are computed and del Prete et al. \cite{del_prete_zero_2018} which compute the zero-step capturable state under arbitrary contact conditions. We apply a dynamic stability model ($\mathbf{a}_1=\mathbf{0}$, $\mathbf{a_2}\neq\mathbf{0}$) in our planner, using a formulation introduced in the following section.
\section{Stability Modeling}
\begin{figure}[t]
    \centering
    \includegraphics[trim={0.8cm 0 0.1cm 0cm},clip,width=\columnwidth]{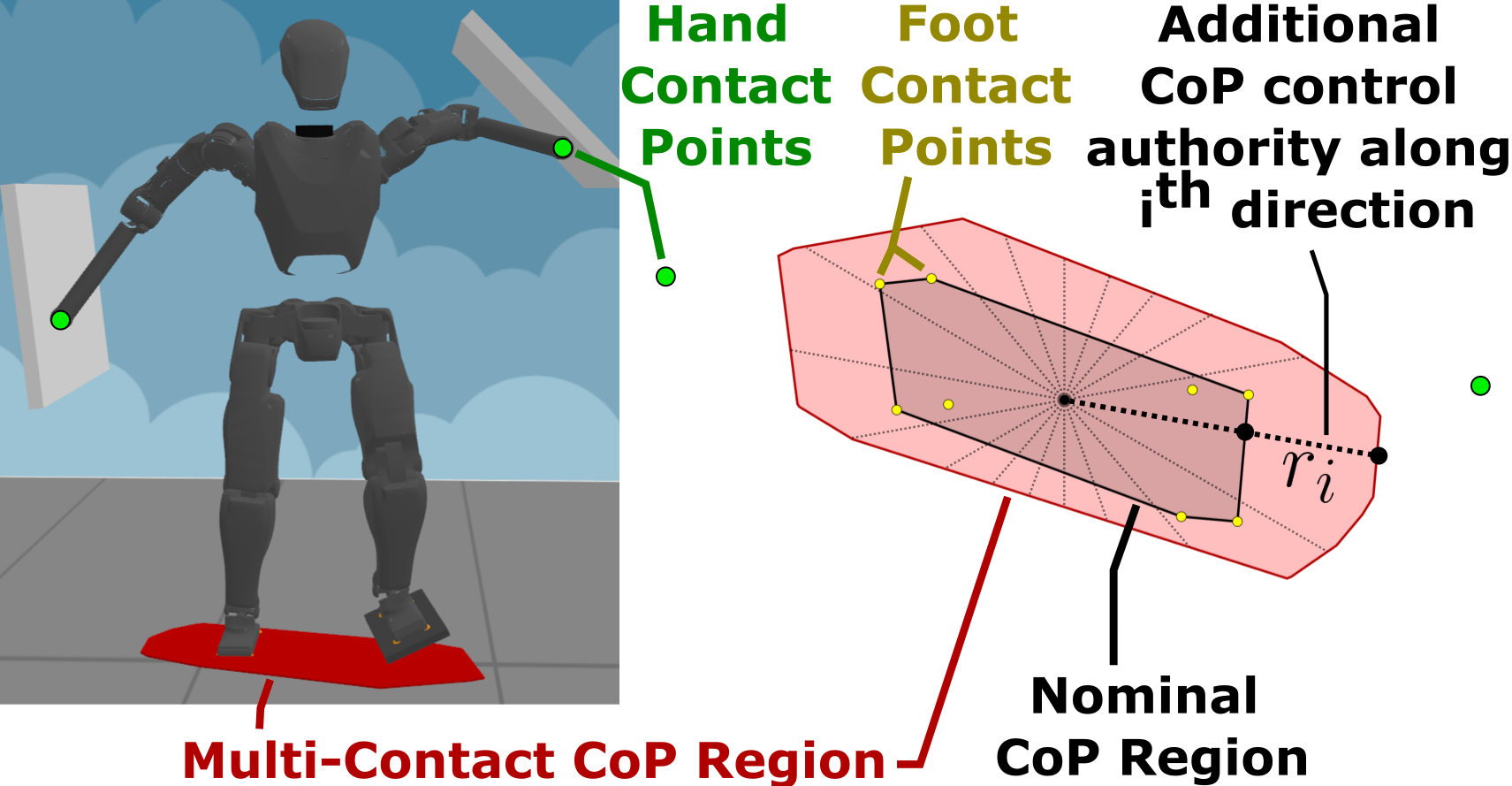}
    \caption{(Left) Example multi-contact scenario in which the robot has two feet and two hand contacts. The robot can feasibly place the CoP in the red convex region, which is computed through successive evaluations of Eq. (\ref{eq:CoPOptimization}). (Right) We train a neural network to learn the $r_i$, the additional CoP control authority along query direction $i$ given by the hand contacts.}
    \label{fig:stability_model_example_graphic}
    \vspace{-4mm}
\end{figure}

In this section, we introduce our method for modeling the robot's feasible CoP through classical optimization and supervised learning. We refer to the robot's \textit{nominal CoP region} as the convex hull of the supporting feet and the \textit{multi-contact CoP region} as the region where the CoP can be placed in the presence of hand contacts, which we compute in the following subsection. These two regions are depicted in Fig. \ref{fig:stability_model_example_graphic}.

\subsection{Optimization-based stability model}\label{sec:optimization-approach}
We begin by using Eq. (\ref{eq:ContactFeasibility}) to model the robot's feasible CoP, adopting a modification of \cite{audren_3-d_2018} to model an LIPM. The following LP optimizes the $xy$ centroidal force $\mathbf{f}_{xy}$ along a query direction $\mathbf{a} \in \mathbb{R}^2$,

\begin{equation} \label{eq:CoPOptimization}
\begin{aligned}
    \max_{\mathbf{f}} \qquad & \mathbf{a}^T\mathbf{f}_{xy} \\
    \textrm{s.t.} \qquad & \sum_k \, f_{z,k} = m ||\mathbf{g}|| \\ 
    & \sum_k \mathbf{p}_k \times \mathbf{f}_k = -\mathbf{c} \times (m\mathbf{g}) \\
    & \mathbf{Bf} \leq \mathbf{0} \\
    & \boldsymbol{\tau}^{-} \leq \mathbf{g}(\mathbf{q}) - \mathbf{J}(\mathbf{q})^T_c\mathbf{f} \leq \boldsymbol{\tau}^{+} \\
\end{aligned}
\end{equation}
for a robot of mass $m$, where contact point $k$ has position $\mathbf{p}_k$ and force $\mathbf{f}_k$, $\mathbf{g}(\mathbf{q})$ is the vector of gravitationally induced torques and $\mathbf{Bf}\leq\mathbf{0}$ is a linearized friction cone. The optimal force $\mathbf{f}^\ast$ is then mapped to a corresponding CoP, $\mathbf{x}_{cop}^\ast$, using an LIPM model relationship
\begin{equation}
    \mathbf{x}^\ast_{cop} = \mathbf{c}_{xy} - \frac{z_0}{mg}\mathbf{f}^\ast_{xy},
\end{equation}
where $z_0$ is the robot's nominal CoM height and $\mathbf{c}_{xy}$ is the 2d CoM position. Vertices of the multi-contact CoP region are computed by successively solving Eq. (\ref{eq:CoPOptimization}) and varying the query direction $\mathbf{a}_i$. Fig. \ref{fig:stability_model_example_graphic} shows the robot in an example contact state and the resulting CoP region (in red).



\begin{figure}
    \centering
    \includegraphics[width=\columnwidth]{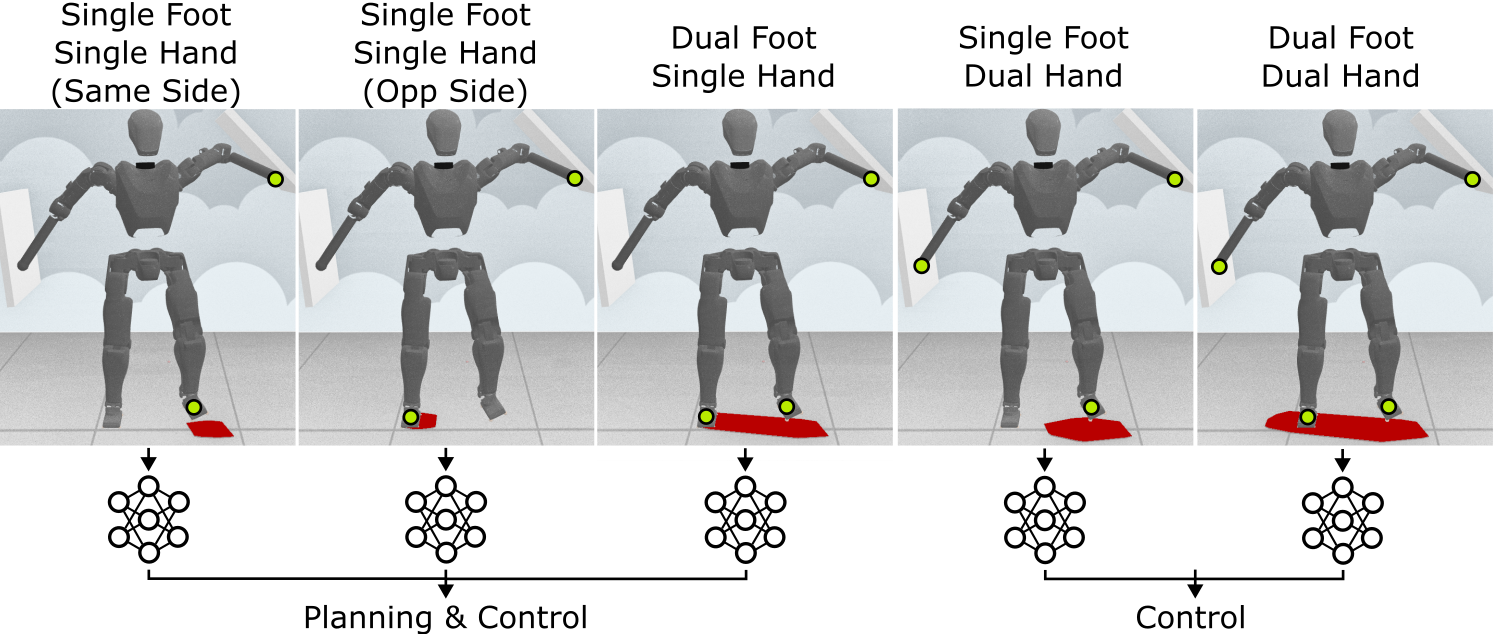}
    \caption{Depiction of the 5 modeled contact permutations for mixed hand and feet contacts, in which at least 1 foot and 1 hand are in contact. The resulting CoP region is shown in red for each example configuration. We train a distinct network to learn the CoP region of each contact configuration.}
    \label{fig:contact-permutations}
    \vspace{-4mm}
\end{figure}

\subsection{Feasible CoP region neural networks}
This work aims to achieve reactive hand contact planning and real-time control to enable a humanoid to brace in the presence of a disturbance. The optimization-based calculation of the feasible CoP region described in Sec. \ref{sec:optimization-approach} has a number of drawbacks for use in a time-critical context. Notably, it is parameterized by whole-body configuration $\mathbf{q}$, requiring Inverse-Kinematics to preview the multi-contact CoP region, which is computationally expensive. Computing the contact Jacobian SVD to accurately model singularities is additional overhead. Finally, computing the multi-contact CoP region requires iteratively solving Eq. (\ref{eq:CoPOptimization}) along multiple query directions.

To reduce computational overhead during planning and control, we train a set of neural networks which model the multi-contact CoP region, given the robot's contact state. We assume quasi-flat foot contact surfaces, meaning the robot can place the CoP anywhere in the nominal CoP region (i.e. the convex hull of the supporting foot/feet). Therefore, the nominal CoP region is a subset of the multi-contact CoP region. For this reason, the neural network predicts a vector $\mathbf{r} = (r_0 \ldots r_n)^T$, which encodes additional CoP control authority beyond the nominal CoP region. As shown in Fig. \ref{fig:stability_model_example_graphic}, an entry of this vector, $r_i$, is the distance from the nominal region to the multi-contact region along a ray originating from the robot's mid-feet in query direction $i$. We train our network along a fixed number of $n=18$ equally spaced query directions. 

\renewcommand{\arraystretch}{1.05}
\begin{figure}[t]
    \centering
    \begin{minipage}[t]{0.36\columnwidth}
        \vspace{0pt}
        \centering
        \scriptsize
        \setlength{\tabcolsep}{3pt}
        \begin{tabular}{ |c c | c c | c c c |}
            \hline
            \rotatebox[origin=c]{90}{Single Hand} & \rotatebox[origin=c]{90}{Dual Hand} & \rotatebox[origin=c]{90}{Single Foot} & \rotatebox[origin=c]{90}{\;Dual Foot\;} & \rotatebox[origin=c]{90}{Contact dim.} &
            \rotatebox[origin=c]{90}{CoM + Stance dim.} &
            \rotatebox[origin=c]{90}{Total input dim.} \\
            \hline
            \checkmark & & \checkmark & & 6 & 3 & 9 \\
            $\checkmark^\ast$ & & \checkmark & & 6 & 3 & 9 \\
            \checkmark & & & \checkmark & 6 & 6 & 12 \\
            & \checkmark & \checkmark & & 12 & 3 & 15 \\
            & \checkmark & & \checkmark & 12 & 6 & 18 \\
            \hline
        \end{tabular}
    \end{minipage}
    \hfill
    \begin{minipage}[t]{0.61\columnwidth}
        \vspace{6pt}
        \centering
        \includegraphics[width=\linewidth]{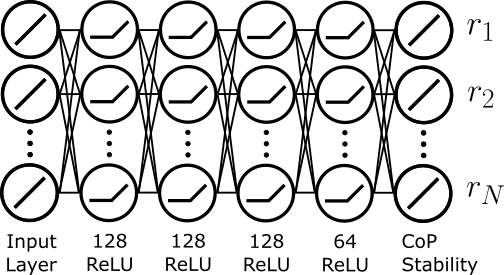}
    \end{minipage}
    \caption{Left: Contact states considered in this work their feature dimension, $^\ast$note there are two single-handed, single-foot states to cover contact on the same and opposites sides as the stance foot. Right: Neural network architecture which computes the vector $\mathbf{r}$, describing the additional CoP control given by the hand contacts.}
    \label{fig:stability-network}
    \vspace{-4mm}
\end{figure}

\begin{figure*}[t]
    \centering
    \begin{subfigure}{0.48\textwidth}
        \includegraphics[width=0.95\textwidth]{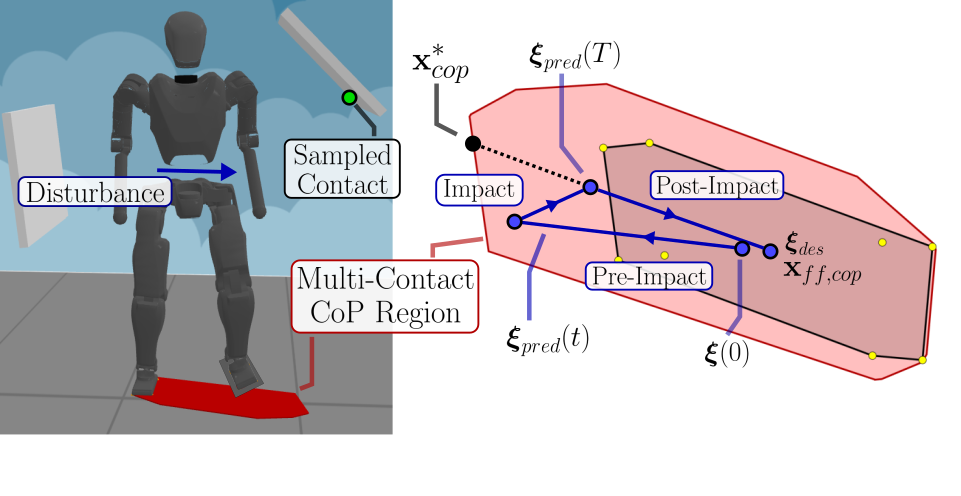}
        \vspace{-4mm}
        \caption{Recovery model: Standing}
    \end{subfigure}
    \begin{subfigure}{0.48\textwidth}
        \includegraphics[width=0.95\textwidth]{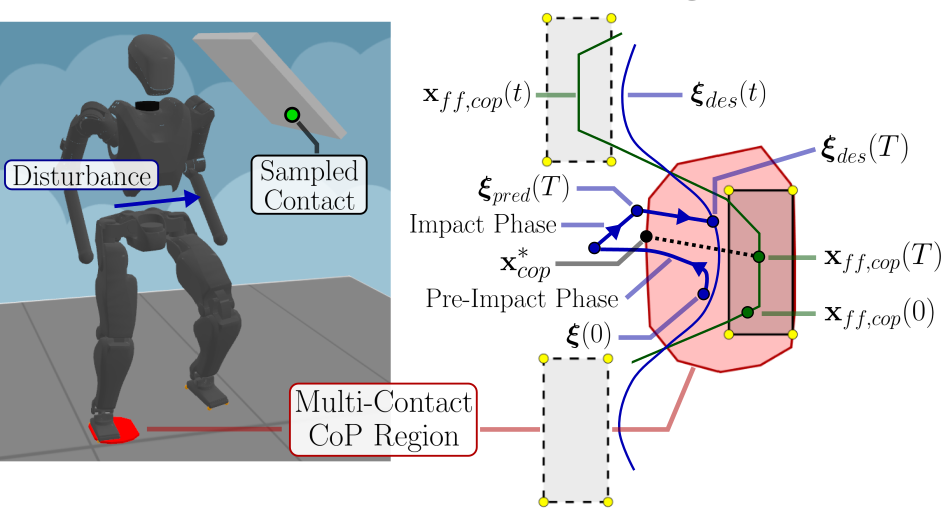}
        \vspace{-4mm}
        \caption{Recovery model: Walking}
    \end{subfigure}
    \caption{Centroidal-based evaluation of a sampled contact point. The predicted capture point trajectory $\mathbf{\xi}_{pred}$ is propagated, starting at the time of planning $t=0$, through impact, $t=T$. The feasible multi-contact CoP region is used to compute an optimal post-impact CoP, $\textbf{x}^\ast$. The contact point's score $s$ is computed to represent the CoP control authority at post-impact.}
    \label{fig:capture-point-rollout}
    \vspace{-4mm}
\end{figure*}
We train a distinct network to model the CoP region of each permutation of hand and feet contacts, in which there is at least 1 hand and 1 foot in contact. Although there are a total of 9 such contact permutations (3 lower body $\times$ 3 upper body), this includes 4 pairs of mirrored contact configurations. Since all properties of \doubleBlind{the}{the Alex} robot are symmetric across the sagittal plane, the mapping of contacts to stability region is also symmetric across the sagittal plane. Therefore, we only need to consider one of each mirrored contact pairing. This leaves 5 distinct (un-mirrored) contact configurations, shown in Fig. \ref{fig:contact-permutations}.

The input and output of the network are expressed locally in mid-feet frame, which has an origin and yaw matching the average of both feet in double support or the supporting foot in single support. The network input includes the contact positions and normals of the hands (6 entries per contact). The input also includes the 3d CoM position and the robot's stance width, length and relative foot yaw if in double support. Fig. \ref{fig:stability-network} (left) shows the networks' dimensionality.

We train a network for each contact configuration using PyTorch \cite{paszke2019pytorch}, with example data generated by sampling random robot configurations and computing the optimization-based region. The network is trained using mini-batch updates with the Adam optimizer to minimize mean squared error, while validation performance is evaluated at each epoch to monitor generalization and guide model selection. Single hand networks are trained with 50\,k examples and 600 epochs and double hand networks with 100\,k examples and 800 epochs, with a learning rate of $7\times10^{-4}$.



\section{Contact Planning}

In this section, we apply the stability networks from the previous section in a sampling-based hand contact planner. We first introduce the planar region map, followed by a centroidal-based rollout for scoring sampled contacts, then detail how this scoring function is applied in our planner.

\subsection{Planar Region Mapping}

\doubleBlind{Our}{The Alex} humanoid is equipped with a ZED X Mini depth camera on the head. This depth image is used to extract a set of convex \textit{planar regions}, which serves as the robot's map of bracing surfaces. Planar region extraction is GPU-accelerated and streams at 30\,Hz, in the style of \cite{grandia_doc_2023, mishra2021gpu}. To prevent contact near the edge of surface, regions are shrunk by a safety distance of 7\,cm.

\subsection{Centroidal Dynamics Rollout}\label{sec:scoring-function}
Candidate contacts are evaluated by rolling out the robot's centroidal dynamics through pre-impact, impact and post-impact phases. In aiming for fast ($\leq$10\,ms), reflexive planning of hand contacts, we take two simplifying assumptions. (i) Hand contacts are planned separately per-side, with the assumption that often the optimal strategy for a given side is independent of the other side. (ii) During rollout, we chose to ignore recovery strategies other than bracing with the upper body and using the multi-contact CoP region. This includes step adjustment, upper-body angular momentum and CoP feedback within the nominal CoP region. Note that these strategies are used by the controller. However, we observe these strategies do not conflict with the use of hand contacts and add significant complexity to previewing the robot's state at the planning phase.

Control of the robot's capture point, $\boldsymbol{\xi}$, is characterized by the following trajectories\doubleBlind{}{ \cite{seydeInclusion2018}}.
\begin{itemize}
    \item $\boldsymbol{\xi}_{des}(t)\,$ desired capture point trajectory
    \item $\textbf{x}_{ff,cop}(t)\,$ nominal, feed-forward CoP trajectory
\end{itemize}
We roll out the capture point dynamics by computing a predicted capture point trajectory, $\boldsymbol{\xi}_{pred}(t)$, through the predicted time of impact $T$. This predicted impact time is computed assuming a maximum linear hand acceleration, $a_{max}$. The predicted capture point is initialized to the robot's current capture point $\boldsymbol{\xi}$ at planning time ($t=0$). Fig. \ref{fig:capture-point-rollout} shows an example capture point rollout for standing and walking.

\textit{Pre-Impact:} The pre-impact phase is evaluated by integrating the first-order capture point dynamics over a duration $T$ required for the hand to established contact.
\begin{equation}\label{eq:capturePointDynamics}
    \dot{\boldsymbol{\xi}} =  \omega (\boldsymbol{\xi} - \textbf{x}_{ff,cop}(t))
\end{equation}
where $\omega=\sqrt{g/\Delta z_{com}}$ is the robot's natural frequency and $\textbf{x}_{ff,cop}(t)$ is the robot's nominal CoP without feedback (e.g. mid-foot when standing).

\begin{figure}
    \centering
    \includegraphics[width=\columnwidth]{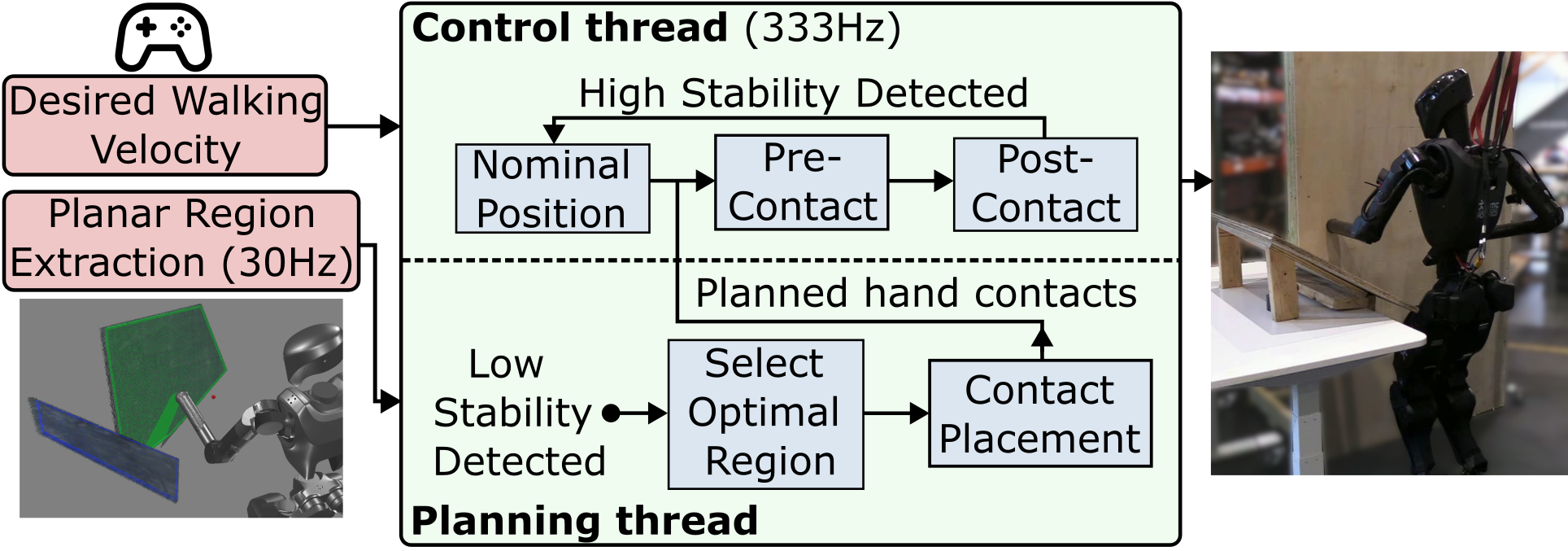}
    \caption{Control flow of our reactive bracing framework. Candidate bracing planar regions are extracted from a head-mounted depth camera. The planning thread is triggered if sufficiently high capture point tracking is detected, which first computes an optimal bracing region per-side, by sampling each region once, and then computing an optimal contact point within the region, by sampling multiple times in the region. The planner dispatches the plan to the controller, which makes contact until sufficiently high stability is regained.}
    \label{fig:control-flow}
    \vspace{-3mm}
\end{figure}

\textit{Impact:} We model impact impulse of fixed magnitude $\Delta p$ imparted along the normal $\mathbf{n}$ of the bracing surface. This causes a jump of the capture point given by
\begin{equation}
\boldsymbol{\xi}^+ = \boldsymbol{\xi}^- + \frac{\Delta p}{m\omega}\textbf{n}_{xy},
\end{equation}
where $\boldsymbol{\xi}^{-/+}$ is the capture point before/after impact, $m$ is the robot's mass and $\mathbf{n}_{xy}$ is the $xy$ component of the surface normal. Through initial simulation tests, we observed $\Delta p = 4 Ns$ as a good approximation of the hand contact impulse, which correlates to a 1.6cm capture point displacement for \doubleBlind{our humanoid}{Alex}.

\textit{Post-Impact \& Recovery Score:} The robot's post-impact balance state is assessed using the multi-contact CoP introduced in the previous section, by querying the network corresponding to the robot's contact state at the predicted time of impact. While it would be possible to continue rolling out the centroidal dynamics using the capture point's linear dynamics (Eq. (\ref{eq:capturePointDynamics})), this would require evaluating the neural network each integration step to update the feasible CoP region. To avoid this additional overhead, we instead consider a snapshot of the CoP control authority at post-impact. This is characterized by the predicted error at post-impact, given by
\begin{equation}\label{eq:error-capture-point}
    \boldsymbol{\xi}_{err} = \boldsymbol{\xi}_{pred} - \boldsymbol{\xi}_{des},
\end{equation}
and corresponding CoP feedback,
\begin{equation}\label{eq:feedback-cop}
    \textbf{x}^\ast_{cop} = \textbf{x}_{ff,cop} + \textbf{x}_{fb,cop}^\ast.
\end{equation}
The optimal post-impact CoP, $\textbf{x}^\ast_{cop}$, is computed by projecting the feedforward CoP, $\textbf{x}_{ff,cop}$, to the boundary of the multi-contact CoP, along the direction of error $\boldsymbol{\xi}_{err}(T)$ (see Fig. \ref{fig:capture-point-rollout}). The vector $\mathbf{x}_{fb,cop}^\ast$ designates the feedback offset. We assign score $s$ to the contact point according to the CoP control authority.
\begin{equation}
    s = (\textbf{x}^\ast_{fb,cop} - \boldsymbol{\xi}_{err}(T)) \cdot \hat{\boldsymbol{\xi}}_{err}(T)
\end{equation}
This score is motivated by the error dynamics of the capture point and serves as an approximation of the post-impact recovery time (see Appendix for details).

\subsection{Select Optimal Region (1 sample per region per side)}
Contact planning is triggered based on capture point tracking error $\boldsymbol{\xi}_{err}$. The planning thread is triggered under the following conditions.
\begin{equation}
\begin{aligned}
	\text{trigger planner} = \begin{cases}
        |\boldsymbol{\xi}_{err}| \geq 3.0\,\text{cm} & \text{if standing} \\ 
        |\boldsymbol{\xi}_{err}| \geq 4.5\,\text{cm} & \text{if walking} \\ 
        \Delta t \geq t_b
    \end{cases}
\end{aligned}
\end{equation}
where $\Delta t$ is the time since last planning and $t_b=3$\,s is a minimum time between plans.

\begin{figure}[t]
    \centering
    \includegraphics[width=\columnwidth]{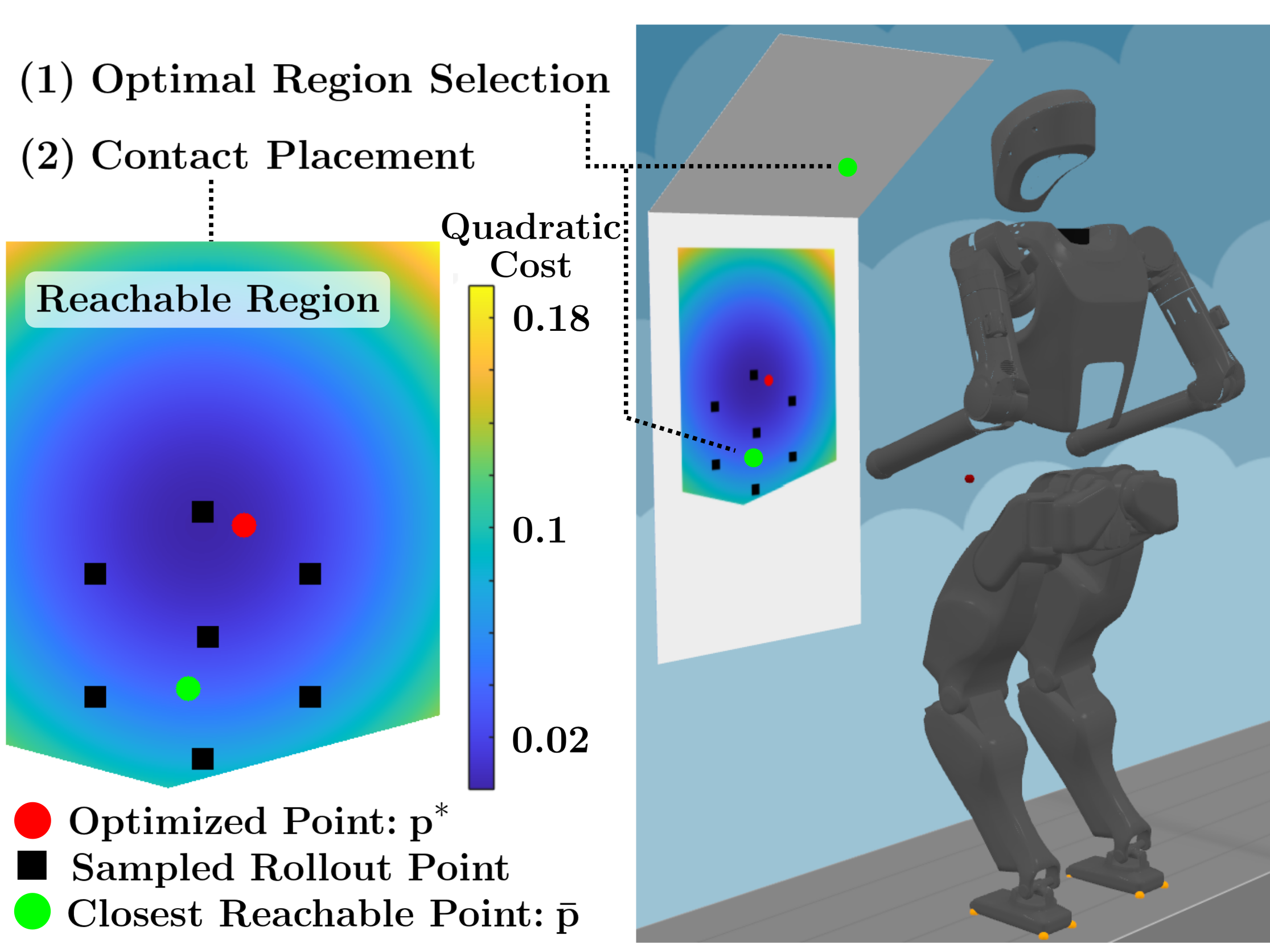}
    \vspace{-3mm}
    \caption{Points in the optimal bracing surface are sampled and scored using a centroidal rollout (Sec. \ref{sec:scoring-function}). The optimal brace point $\mathbf{p}^\ast$ is computed through a QP that models the score gradient and regularizes on the closest reachable point (Sec. \ref{sec:contact_placement}). Here, the region color depicts the cost function of the QP.}
    \label{fig:recovery_graphic}
    \vspace{-3mm}
\end{figure}

Once triggered, the planner iterates through all pairs $(i,R)$ for side $i\in\{\text{left,right}\}$ and region $R$ to select an optimal bracing region for each hand. For each pair $(i,R)$, the candidate region is cropped to represent the reachable sub-portion of the region (see Appendix for details on reachability model). If the region is reachable, the planner then computes a score $s_{iR}$ for the closest reachable point $\bar{\mathbf{p}}_{iR}$ on the region, using the centroidal rollout discussed in Sec. \ref{sec:scoring-function}. The region with the highest score (if present) is selected as the bracing surface for side $i$.

\subsection{Contact Placement (n samples per side)}\label{sec:contact_placement}
Once an optimal bracing surface $R^*$ is selected (if one is present), contacts are sampled within the surface to determine an optimal contact position $\mathbf{p}^\ast$ (we omit subscript notation for brevity, all variables in this subsection are with respect to side $i$ and optimal region $R^\ast$). These samples are centered on the closest reachable point $\bar{\mathbf{p}}$, and $n=6$ points sampled radially around it at a radius $r_s=12$cm. If $\bar{\mathbf{p}}$ is near the boundary of the reachable workspace, samples are shifted inwards such that the center is a distance $r_s$ from the edge.

The gradient of the score $s$ within the region is modeled as a linear function of a point $\mathbf{p} \in \mathbb{R}^2$ in the region, expressed in local coordinates,
\begin{equation}
    \hat{s}(\mathbf{p}) = \mathbf{a}^T\mathbf{p} + b
\end{equation}
for some $\mathbf{a}\in \mathbb{R}^2$, $b\in \mathbb{R}$, which are computed through a least-squares fit
\begin{equation}
    (\mathbf{a}^\ast, b^\ast) = \argmin_{\mathbf{a},b} \sum_i (\mathbf{a}^T\mathbf{p}_i + b - s_i)^2
\end{equation}
where $s_i$ is the score of the $i^{th}$ sampled point. We compute the optimal contact placement $\mathbf{p}^\ast$ within the region by solving the quadratic program
\begin{equation}
\begin{aligned}    
    \mathbf{p}^\ast = \argmin_{\mathbf{p}} \,\, & w |\mathbf{p} - \bar{\mathbf{p}}|^2 + \mathbf{a}^{\ast T}\mathbf{p} \\
    \text{s.t.} \quad & \mathbf{C}\mathbf{p} \leq \mathbf{d} \\
\end{aligned}
\end{equation}
where $\mathbf{C},\mathbf{d}$ encodes the reachable portion of the region (see Appendix). This biases contact placement towards the closest reachable point while adjusting based on the computed scores, in order to increase the CoP control authority of the planned contact at post-impact.

\section{Bracing Control}

This section describes our control method for realizing the planned hand contacts. Similarly to the planner, the control is divided into pre-impact and post-impact phases.

\subsection{Pre-Impact Bracing Control}
During the pre-contact phase, the hand follows a cubic polynomial taskspace trajectory expressed in world-frame, computed through the following conditions.
\begin{itemize}
    \item Initial position $\mathbf{p}_0$, velocity $\mathbf{v}_0$ match hand's current state,
    \item Final position $\mathbf{p}_f$ is the planned contact point,
    \item Final velocity $\mathbf{v}_f = -v_f\mathbf{n}$,
    \item Trajectory duration: $T = \sqrt{|\mathbf{p}_f - \mathbf{p}_0|\, / \,a_{max}}$,
\end{itemize}
where $v_f=0.9\,\text{m/s}$ is a fixed touchdown velocity, $a_{max} = 9\,\text{m/s}^2$ is an approximation of hand's maximum linear acceleration, and $\mathbf{n}$ is the surface normal. Touchdown is triggered based on the following conditions of the hand state, expressed in the frame of the planar region ($xy$ along plane, $z$ points along the surface normal).
\begin{itemize}
    \item $\mathbf{p}_{xy} \in \mathbb{P}$, the hand is inside the region polygon $\mathbb{P}$,
    \item $p_z \leq \Delta z_{max}$, the hand is sufficiently close to the region,
    \item $v_z \leq v_{z,max}$, the hand has sufficiently low speed,
\end{itemize}
where $z_{max}=4\,\text{cm}$, to account for error in the estimated plane and the estimated robot position and $v_{z,max}=30\,\text{cm/s}$.

\subsection{Post-Impact Bracing Control}
After impact, a set of inverse-dynamics control objectives are generated. A zero-acceleration objective along the planar region's $z$-axis (surface normal) is created. Along the $xy$-plane, either a PD position feedback or zero-acceleration objective is used, depending on the desired arm load force ($\mathbf{f}_{des}$) from the previous control tick, to eliminate drift if the hand is lightly loaded.
\begin{equation}
\begin{aligned}    
    \mathbf{J}_{xy} \ddot{\mathbf{q}} = \begin{cases}
        \text{PD}(\mathbf{p}_{cur},\mathbf{p}_{des}) & |\mathbf{f}_{des}| \leq 12\,\text{N} \\
        \qquad \mathbf{0} & \text{else}\; (\text{sufficiently loaded})
    \end{cases}
\end{aligned}
\end{equation}
Every control tick the controller runs the neural network corresponding to the current contact conditions. This in turn allows the controller to place the desired CoP at the maximal, feasible position for recovery, given the robot's current state.

Contact is released and the arm returns to holding nominal position under the following conditions.
\begin{itemize}
    \item $|\boldsymbol{\xi}_{err}| \leq \{ 1.5\,\text{cm (standing), } 2.5\,\text{cm (walking)}\}$,
    \item $\boldsymbol{\xi}$ is 5\,cm inside nominal CoP region,
    \item The hand deviates 9\,cm from the touchdown position,
    \item The hand is near the edge of the reachable workspace.
\end{itemize}

\section{Results}


\subsection{Neural network evaluation}
Model prediction accuracy is evaluated using the root mean squared error (RMSE) between the network outputs and example data computed through optimization (Eq. (\ref{eq:CoPOptimization})). Let $\hat{\mathbf{r}} \in \mathbb{R}^{D}$ denote the predicted output vector, $\mathbf{r} \in \mathbb{R}^{D}$ the corresponding ground truth and $D=18$ the output dimension. The RMSE, $e_{rmse}$, is then computed as 
\begin{equation}
e_{rmse} = \sqrt{\frac{1}{N D} \sum_{i=1}^{N} |\hat{\mathbf{r}}_i - \mathbf{r}_i|^2},    
\end{equation}
where $N$ is the number of evaluation samples. Tab. \ref{table:nn_evaluation} shows the RMSE errors, computed from an evaluation dataset that is distinct from the training dataset. Note that this error is mitigated by our modeling choice, such that erroneous negative entries of $\hat{\mathbf{r}}$ are ignored during runtime, as the multi-contact CoP region cannot under-approximate the nominal CoP region by design. Furthermore, we do expect non-zero error, as the network is only receiving partial information, as the ground truth value is characterized by the whole-body configuration $\mathbf{q}$ of the robot.

\begin{table}[t]
\begin{center}
\caption{Neural Network Evaluation}\label{table:nn_evaluation}
\vspace{-2mm}
\begin{tabular}{ | @{\hspace{4pt}} c @{\hspace{4pt}} | @{\hspace{4pt}} c @{\hspace{4pt}} | } 
\hline
 Contact config. & $e_{rmse}$ [cm] \\
\hline\hline
Single foot, single hand (same side) & 1.36 \\
\hline
Single foot, single hand (opposite side) & 1.06 \\
\hline
Dual foot, single hand & 0.89 \\
\hline
Single foot, dual hand & 2.21 \\
\hline
Dual foot, dual hand & 2.62 \\
\hline
\end{tabular}
\end{center}
\vspace{-4mm}
\end{table}

\subsection{Simulation}

\begin{figure}[t]
    \centering
    \includegraphics[width=\columnwidth]{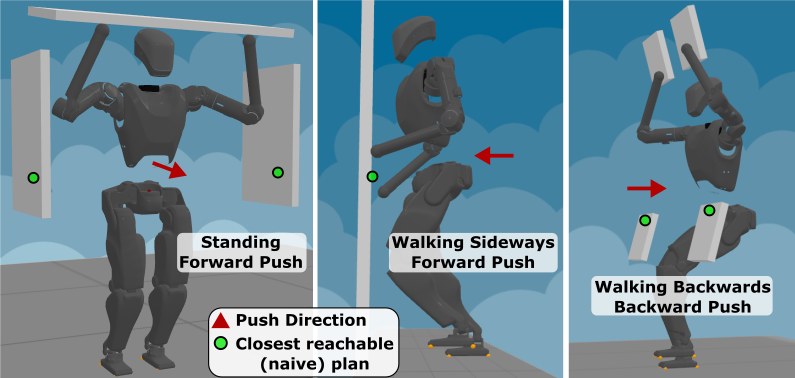}
    \vspace{-4mm}
    \caption{Three scenarios were tested in simulation. In each, we determine the maximum sustainable impulse for (i) no bracing, (ii) naive (closest reachable) bracing, and (iii) bracing using the presented planner.}
    \vspace{-4mm}
    \label{fig:simulation_cases}
\end{figure}

We tested our framework in simulation to determine the maximum sustainable impulse in three scenarios. Shown in Fig.~\ref{fig:simulation_cases}, these include standing, walking sideways, and walking backwards. At a fixed time in the simulation, an impulse is applied to the pelvis link. We perform a binary search to determine the maximum impulse that can be sustained without the robot falling. To quantify the effectiveness of our approach, we compare against two baselines: (i) no bracing and (ii) a naive planner that braces using the closest reachable point. In Fig.~\ref{fig:simulation_cases}, the naive contact plans are depicted. For the standing and backwards walking simulations, the naive plan selects different bracing regions, which may be difficult to determine using simple heuristics. The sideways walking simulation only has a single bracing surface available, which is reachable by the left hand. This highlights the value of contact placement, which adjusts the contact by 14.9\,cm, yet it yields 16\,$\%$ improvement in impulse resilience.

\begin{table}[t]
\begin{center}
\caption{Maximum Sustainable Impulse [N$\cdot$s]}\label{table:simulation-push-resilience}
\vspace{-2mm}
\begin{tabular}{ |c|c|c|c| } 
\hline
 Scenario & No & Naive & Optimized \\
 & Bracing & Bracing & Bracing \\
\hline\hline
Standing, For. Push & 15.6 & 25.8 & 29.4 \\
\hline
Walk Sideways, For. Push & 17.7 & 36.4 & 42.3 \\
\hline
Walk Backwards, For. Push & 23.0 & 26.6 & 32.1 \\
\hline
\end{tabular}
\end{center}
\vspace{-2mm}
\end{table}

\begin{table}[t]
\begin{center}
\caption{Planner Benchmark}\label{table:timing_evaluation}
\begin{tabular}{ | c | c | c | } 
\hline
 & Numerical  & Neural  \\
 &  Optimization &  Network \\
\hline\hline
CoP Region (per rollout) & 0.65\,ms & 0.13\,ms \\
\hline
IK (per rollout) & 13\,ms & - \\
\hline
Additional computation (per side) & \multicolumn{2}{c|}{3\,ms} \\
\hline
\hline
Total (2 sides, 14 rollouts/side) & \textbf{388}\,ms & \textbf{9.7}\,ms \\
\hline
\end{tabular}
\end{center}
\vspace{-3mm}
\end{table}

\begin{figure*}[t]
    \centering
    \includegraphics[width=\textwidth, keepaspectratio]{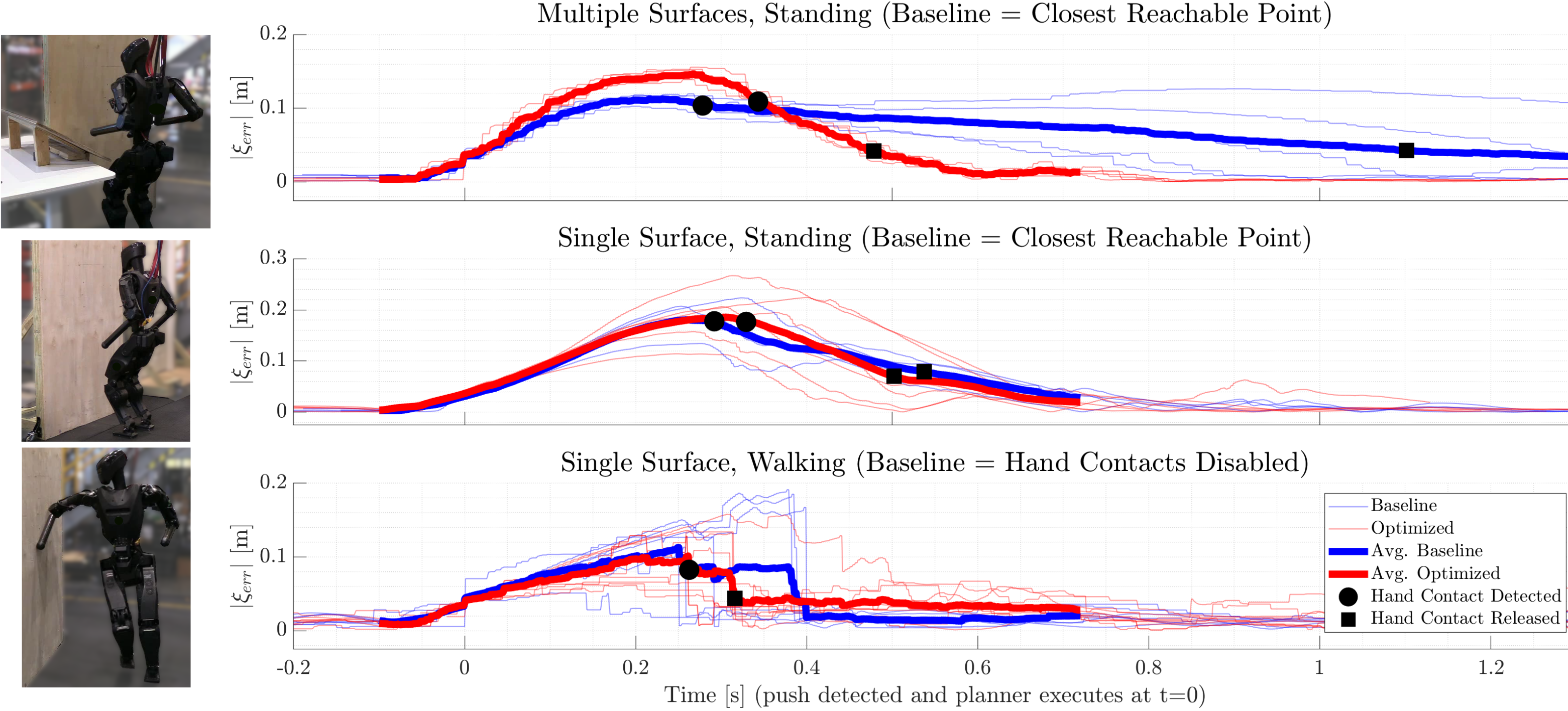}
        \vspace{-2mm}
    \caption{We tested our planner in the scenario shown, where the robot has two available regions to brace with the left hand: a vertical wall in front and a slanted surface on the side. The solid markers indicate the average time and corresponding capture point error when impact is detected.}
    \label{fig:recovery-plot-results}
    \vspace{-4mm}
\end{figure*}

\subsection{Hardware}
We conducted three rounds of push recovery tests to validate our framework. We alternate each push between a baseline and our presented (optimized) planner. The resulting error plots and their average is plotted in Fig.~\ref{fig:recovery-plot-results}. While push magnitude cannot be directly measured, we consider the average slope of the capture point error at push time as a proxy for push magnitude, which had little ($<5\%$) variation between baseline and optimized datasets.

\textit{(1) Standing, Multi-Surface (n=10 pushes)}: in this experiment, the robot is standing with a wall in front and tilted surface on the side. The baseline plan is to select the closest reachable brace point, which is on the tilted surface, while the optimized plan reaches for the wall. As shown in Fig.~\ref{fig:recovery-plot-results}, the wall yields larger tracking error in part due to longer time to contact. We observe a substantial (57\,\%) decrease in average recovery time for the optimized plan.

\textit{(2) Standing, Single Surface (n=10 pushes)}: the robot is facing a wall, and the planner ignores the right hand in order to test single-handed bracing. The optimized plan adjusts the contact point upwards along the wall by approximately 12\,cm relative to the baseline plan. This relatively small adjustment decreases the average post-impact recovery time by 29\%, indicating a higher CoP control authority.

\textit{(3) Walking, Single Surface (n=12 pushes)}: in this experiment, the robot is commanded to walk back and forth alongside a wall, switching between forward and sideways gaits. The robot is pushed with randomized timing and direction. The baseline for this experiment is the pushes that do not trigger the contact planner, and rely only on lower body recovery strategies. On average, the robot achieves high stability and releases contact 316\,ms after the push is detected. In the baseline trials (no hand contacts), high stability is achieved slightly later (384\,ms), meaning the bracing strategy yields an 18\,\% reduction in recovery time. As seen in Fig.~\ref{fig:recovery-plot-results}, the average capture point error remains relatively high ($\approx$3\,cm) after contact is released, until $\approx$750\,ms after push time when it converges with the average baseline error. We attribute this to the momentum of the arm's motion upon releasing contact, which is not modeled in the nominal centroidal plan. We also observe that the duration of hand contact is short ($\approx55\,$ms) compared to the standing tests, such that the impact is sufficient to stabilize the robot. We attribute this partly to latency in contact detection, such that the post-impact phase is abbreviated. Subsequent tests will aim to test higher push forces, which will induce longer post-impact recovery times.




\section{Conclusion}

Bracing with the upper body is a core recovery strategy used when humans are in low stability scenarios \cite{rueckert2016probabilistic}. It is crucial for humanoids to also posses this ability in order to maximize reliability and reduce falls. Towards this goal, we presented a reactive, sampling-based planning approach that is motivated by the preview-based recovery approach from Mason et al. \cite{mason18mpc} and the optimal region selection from Corbères et al. \cite{corberes2025perceptive}. We began by approximating the optimization-based approach to compute feasible CoP placement by a set of neural networks, by using distinct networks for each contact permutation. These networks allow us to bypass both the Inverse Kinematics and Linear Programming steps required to traditionally compute this region. Our findings show this translates to approximately 50\,$\times$ speedup in computation time. In turn, this allows for rapid assessment of candidate brace points, which we evaluate by rolling out the centroidal dynamics, yielding a total reaction time of $\approx$10 ms. We demonstrate our approach in both simulation and hardware, achieving substantial improvement in stability compared to baseline walking and significant (15-20\,\%) improvement compared to naive bracing strategies.

Our presented approach, though using some data-driven techniques, relies heavily on reduced-order models. Future work could investigate alternative methods for scoring candidate contacts, particularly with respect to the impact map. One of our unexpected findings is that the impact phase, as opposed to the post-impact phase, often accounted for the majority of recovery. Other avenues for incorporating data-driven techniques could explore training networks to directly predict the optimal contacts, given the robot's state and planar region map.

\appendix
\subsection{Reachability Model} \doubleBlind{Our}{The Alex} robot has 3 shoulder joints and 1 elbow joint. We use a spherical model centered on the shoulder with a minimum (32\,cm) and maximum (72\,cm) reachability, centered on the shoulder. We crop the reachable workspace along the sagital plane to prevent the arms colliding. We encode this spherical workspace as a set of linear constraints when cropping the bracing region, in a similar style as \cite{corberes2025perceptive}.
\subsection{Capture point error dynamics} The error dynamics of the capture point are linear, based on the feed-forward trajectories satisfying the LIPM relationship by design.
\begin{equation}
   \dot{\boldsymbol{\xi}}_{err}(t) = \omega (\boldsymbol{\xi}_{err}(t) - \textbf{x}_{fb,cop}(t))
\end{equation}
For a fixed feedback CoP, $\mathbf{x}^\ast_{fb,cop}$, this yields
\begin{equation}
    \boldsymbol{\xi}_{err}(t) = \boldsymbol{\xi}_{err}(T) + (\boldsymbol{\xi}_{err}(T) - \textbf{x}^\ast_{fb,cop}) e^{\omega (t-T)}
\end{equation}
To estimate the time required to reach $\boldsymbol{\xi}_{err} = \mathbf{0}$, assume a fixed CoP feedback offset $\mathbf{x}_{fb,cop} = k \boldsymbol{\xi}_{err}(T)$. This becomes a scalar dynamic relationship along the line $\boldsymbol{\xi}_{err}(0)$, with a recovery time $t_r$ given as
\begin{equation}
    t_r = T + \frac{1}{\omega} \text{ln} \bigg( \, \frac{\xi_{err}(T)}{x^\ast_{fb,cop} - \xi_{err}(T)} \, \bigg),
\end{equation}
where the denominator of the logarithm's argument is equivalent to our scoring function. This means that high scores are correlated with low recovery times. We chose not to score based on $t_r$ in the event that the logarithm's argument is negative. This may occur in cases when the robot is capturable using both hands but not one.

\printbibliography

\balance

\end{document}